\documentclass[11pt]{article}

\usepackage[final]{acl}

\usepackage{times}
\usepackage{latexsym}
\usepackage[T1]{fontenc}
\usepackage[utf8]{inputenc}
\usepackage{microtype}
\usepackage{inconsolata}

\usepackage{graphicx}
\usepackage{subcaption}
\usepackage{booktabs}
\usepackage{longtable}
\usepackage{float}

\usepackage{amsfonts}
\usepackage{nicefrac}

\usepackage{hyperref}
\usepackage{url}
\usepackage{xcolor}

\title{MultiSynt/MT: Trillion-Token Multi-Parallel Pre-Training Data \\Translated Across 36 Languages}

\author{
  \textbf{Maximilian Idahl\textsuperscript{1,2}},
  \textbf{J\"org Tiedemann\textsuperscript{3}},
  \textbf{Sampo Pyysalo\textsuperscript{4}},
  \textbf{David Salinas\textsuperscript{5,6}},
  \textbf{Tomasz Galica\textsuperscript{4}},
\\
  \textbf{Shenbin Qian\textsuperscript{7}},
  \textbf{Tudor Nicolae Mateiu\textsuperscript{8}},
  \textbf{Zihao Li\textsuperscript{3}},
  \textbf{Anna Lokrantz\textsuperscript{9}},
  \textbf{Fedor Vitiugin\textsuperscript{4}},
\\
  \textbf{Andr\'e Martins\textsuperscript{10,11,12}},
  \textbf{Jenna Kanerva\textsuperscript{4}},
  \textbf{Filip Ginter\textsuperscript{4}},
  \textbf{Matthias Lindemann\textsuperscript{10}},
  \textbf{Tim Isbister\textsuperscript{9}},
\\
  \textbf{Birger Mo\"ell\textsuperscript{9}},
  \textbf{Jonas Lindh\textsuperscript{9}},
  \textbf{Jan Haji\v{c}\textsuperscript{13}},
  \textbf{Jenia Jitsev\textsuperscript{14,15,16,17}},
  \textbf{Andrey Kutuzov\textsuperscript{7}},
\\
  \textbf{Stephan Oepen\textsuperscript{7}},
  \textbf{Gema Ram\'irez-S\'anchez\textsuperscript{8}}
\\[0.6em]
  \textsuperscript{1}ellamind \quad
  \textsuperscript{2}Leibniz University Hannover \quad
  \textsuperscript{3}University of Helsinki \quad
  \textsuperscript{4}University of Turku \\
  \textsuperscript{5}ELLIS Institute T\"ubingen \quad
  \textsuperscript{6}Prior Labs \quad
  \textsuperscript{7}University of Oslo \quad
  \textsuperscript{8}Prompsit Language Engineering \\
  \textsuperscript{9}AI Sweden \quad
  \textsuperscript{10}Instituto de Telecomunica\c{c}\~oes \quad
  \textsuperscript{11}Instituto Superior T\'ecnico \quad
  \textsuperscript{12}TransPerfect \\
  \textsuperscript{13}Charles University \quad
  \textsuperscript{14}Ontocord \quad
  \textsuperscript{15}LAION \quad
  \textsuperscript{16}Open-$\Psi$ (Open-Sci) Collective \\
  \textsuperscript{17}Juelich Supercomputing Center (JSC), Research Center Juelich (FZJ)
}

\begin{document}

\maketitle

\begin{abstract}
Open web-scale pre-training corpora remain concentrated in English, limiting multilingual LLM development.
We introduce MultiSynt/MT, an open synthetic parallel corpus with approximately 4.8 trillion target-language tokens across 36 languages, produced by translating 100 billion high-quality Nemotron-CC tokens with {\sc Tower+} and OPUS-MT/HPLT-MT systems.
For many medium- and lower-resource European languages, this is the largest openly available pre-training resource.
Across five high- and medium-resource languages, reference LLMs trained on MultiSynt/MT reach the final score of HPLT 2.0, a native-data baseline, using roughly 72\% fewer pre-training tokens, and outperform it by approximately 15\% relative at a matched 100B-token training budget.
Our analyses also identify evaluation blind spots: standard multiple-choice benchmarks miss translation-quality differences that a fluency-sensitive LLM-as-judge protocol recovers on the trained LLMs without detecting a deficit relative to its native-data baseline, while Norwegian idiomatic and culturally grounded tasks remain better served by native data.
We release the corpus, including row-aligned translations from multiple systems, to support controlled research on multilingual pre-training data and evaluation.
\end{abstract}

\section{Introduction}

Openly available pre-training data at scale exist primarily for English \citep{penedo-etal-2024-fineweb,su-etal-2025-nemotron-cc}, with most other languages lacking volume or quality (see Section~\ref{sec:related} for details).
This shortfall has motivated growing interest in machine translation as a means of producing multilingual pre-training data at scale, but the practice raises legitimate concerns: translated text can carry stylistic artifacts known as ``translationese'' \citep{gellerstam1986translationese,riley-etal-2020-translationese-language} and inherits the cultural reference frame of the source language, with names, places, idioms and culturally grounded knowledge in the English source remaining English-anchored after translation \citep{naous-etal-2024-cultural-bias,yao-etal-2024-cultural-mt}.

\begin{figure*}[t]
    \centering
    \includegraphics[width=\linewidth]{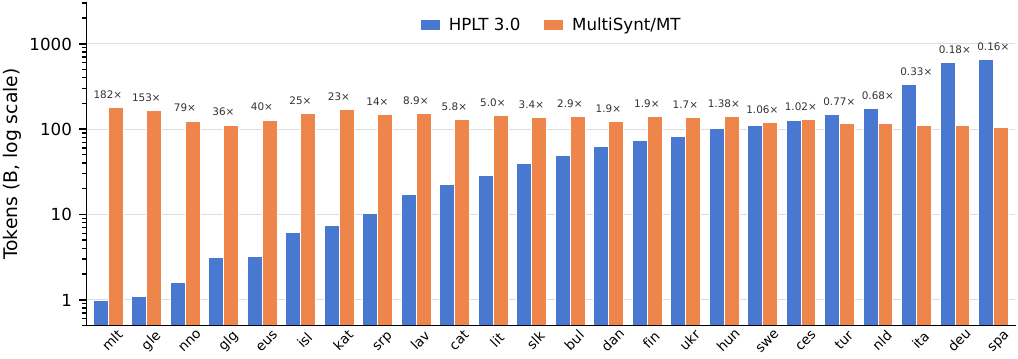}
    \caption{Per-language token counts in MultiSynt/MT and HPLT 3.0 (log scale, Gemma-3 tokenization, 24-language representative subset).
    MultiSynt/MT is larger for 19 of 24 languages, by over an order of magnitude for the lowest-resource ones (e.g.\ mlt 182$\times$, gle 153$\times$, nno 79$\times$); HPLT 3.0 is larger only on deu and spa (5--6$\times$).
    The value above each group gives the MultiSynt/MT-to-HPLT 3.0 token ratio.
    Full statistics are provided in Appendix~\ref{sec:appendix-resource}.}
    \label{fig:resource-comparison}
\end{figure*}

We address this resource limitation by introducing \textbf{MultiSynt/MT}, an open multilingual synthetic parallel corpus of approximately \textbf{4.8 trillion target-language tokens covering 36 languages}, produced by translating a 100B-token sample of high-quality web data from Nemotron-CC with open translation models, including {\sc Tower+} 9B and 72B \citep{rei-etal-2026-tower-plus} and OPUS-MT \citep{tiedemann-etal-2024-opus-mt-democratizing}.
MultiSynt/MT is the largest openly available pre-training corpus to date for many European low- and medium-resource languages, exceeding the largest comparable native resource (HPLT 3.0; \citealp{oepen-etal-2025-hplt-3}) by more than an order of magnitude in the lowest-resource ones (Figure~\ref{fig:resource-comparison}).
For a subset of languages, we release row-aligned parallel translations from multiple systems, enabling controlled comparison under matched source data, and the corpus is released under a permissive open license.

\begin{figure}[!b]
    \centering
    \includegraphics[width=\linewidth]{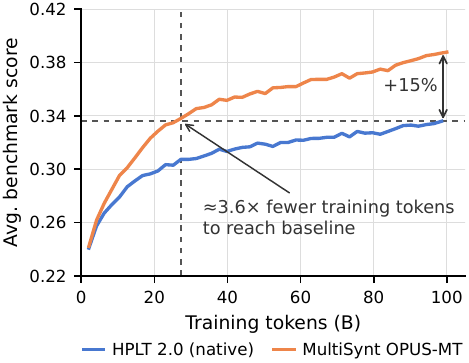}
    \caption{HPLT 2.0 (native) versus MultiSynt/MT (OPUS-MT) on the multilingual benchmark suite, averaged over five languages (dan, nld, ita, por, swe).
    MultiSynt/MT crosses the HPLT 2.0 endpoint at $\sim$28B training tokens and continues to improve through 100B tokens.
    Per-language and per-benchmark breakdowns are provided as supplementary data.}
    \label{fig:gains-compute}
\end{figure}

Alongside the corpus, we present a balanced empirical characterization reporting both where reference LLMs trained on MultiSynt/MT outperform native multilingual baselines and where they fall short, and drawing attention to evaluation practices that obscure these differences.
Across five high- and medium-resource languages, reference LLMs pre-trained on MultiSynt/MT reach the final score of a native-data baseline (HPLT 2.0\footnote{Our pre-training experiments began before the more recent HPLT 3.0 release was available, so we use HPLT 2.0 as the native baseline throughout Section~\ref{sec:effectiveness}.
While the 3.0 version is larger, the distributions of the two are closely similar \citep{oepen-etal-2025-hplt-3}; see also Figure~\ref{fig:noreval}.}), using approximately 72\% fewer training tokens, and improve over it by approximately 15\% relative at a matched 100B-token training budget (Section~\ref{sec:effectiveness}).
A fluency-sensitive LLM-as-judge evaluation on the trained LLMs cleanly recovers an MT-system quality ranking that standard multiple-choice benchmarks fail to discriminate, with no fluency deficit detected relative to the native-data baseline under this protocol (Section~\ref{subsec:fluency-judge}); a complementary embedding-space diagnostic finds that standard benchmark items tend to have more MultiSynt/MT than HPLT 2.0 documents among their nearest neighbors (Section~\ref{subsec:benchmark-alignment}).
On Norwegian tasks targeting idiomatic and locally grounded knowledge, models trained on native data outperform translated-data models in both Bokm{\aa}l and Nynorsk, while localized variants of standard English-origin benchmarks can favor translated data, indicating that the two data sources contribute complementary qualities (Section~\ref{subsec:norwegian}).

\section{Related Work}
\label{sec:related}

\paragraph{Multilingual pre-training corpora.}
Open multilingual pre-training corpora have evolved from filtered Common Crawl pipelines (CCNet, mC4, OSCAR, ROOTS) to larger and better-filtered resources including MADLAD-400, CulturaX, Glot500, HPLT, FineWeb2 and EMMA-style adaptation \citep{wenzek-etal-2020-ccnet, conneau-etal-2020-xlm-r, xue-etal-2021-mt5, abadji-etal-2022-cleaner-corpus, laurencon-etal-2022-roots, bigscienceworkshop-2022-bloom, kudugunta-etal-2023-madlad-400, nguyen-etal-2024-culturax, imanigooghari-etal-2023-glot500, de-gibert-etal-2024-hplt, burchell-etal-2025-hplt, penedo-etal-2025-fineweb-2, ji-etal-2024-emma-500}, but a persistent volume and quality imbalance remains for lower-resource languages \citep{kreutzer-etal-2022-quality-audit}.
MultiSynt/MT targets that part of the language distribution as a complement to these corpora; we use HPLT \citep{oepen-etal-2025-hplt-3}, the strongest openly available naturally occurring multilingual web-text baseline at our scale, as our principal point of comparison.

\paragraph{High-quality English source data.}
Translation-based corpora inherit the quality distribution of their source: recent reference-baseline work shows that Nemotron-CC HQ \citep{su-etal-2025-nemotron-cc} produces the strongest downstream performance across model and token scales, ahead of established English corpora including C4, The Pile, RefinedWeb, FineWeb-Edu and DataComp-LM \citep{raffel-etal-2020-t5, gao-etal-2021-pile, penedo-etal-2023-refinedweb, penedo-etal-2024-fineweb, li-etal-2024-datacomp-lm, nezhurina-etal-2025-open-sci-ref}, motivating its use as the MultiSynt/MT source distribution.

\paragraph{Translated corpora for language-model pre-training.}
Studies on Basque and Indian languages find that machine-translated pre-training data can be competitive with native data on standard NLU \citep{urbizu-etal-2023-mt-rescue,doshi-etal-2024-translationese-pretraining}.
Complementary studies on Arabic, Indonesian, and Tamil document task-specific linguistic or cultural gaps despite useful downstream transfer from translated data \citep{boughorbel-etal-2024-translated-data,velasco-roque-2025-translated-text}.
Closest in scale, TransWeb-Edu / CuatroLLM and its extension TransWebLLM translate up to 1.7T tokens from FineWeb-Edu into nine languages \citep{wang-etal-2024-transwebedu, wang-etal-2025-machine-translated-pretraining}, and the contemporaneous FineTranslations reverses the direction by translating 500+ non-English languages into English \citep{penedo-etal-2026-finetranslations}.
A parallel line studies translation objectives and the inclusion or mixing of bilingual translation data during pre-training or continual pre-training \citep{ji-etal-2024-mt-bridge,li-etal-2024-comparison-objectives,ji-etal-2025-bilingual-translation-data,li-etal-2025-rethinking-multilingual}; we instead train standard next-token language models on translated target-language documents.
MultiSynt/MT extends this line in three ways: 36-language coverage at approximately 4.8T target-language tokens, a Nemotron-CC HQ rather than FineWeb-Edu source, and row-aligned outputs from multiple open MT systems, including translation-specialized LLMs in the Tower family \citep{alves-etal-2024-tower,rei-etal-2026-tower-plus} and classical NMT systems \citep{tiedemann-2012-opus,tiedemann-etal-2024-opus-mt-democratizing,junczys-dowmunt-etal-2018-marian}, supporting controlled comparison under matched source data.

\paragraph{Translationese, cultural grounding, and native-authored evaluation.}
Translated text differs systematically from natively authored text \citep{koppel-ordan-2011-translationese,volansky-etal-2015-translationese-features,vanmassenhove-etal-2021-machine-translationese,bizzoni-etal-2020-machine-translationese}, with downstream effects on cross-lingual benchmark interpretation \citep{graham-etal-2020-statistical-power,freitag-etal-2020-bleu,artetxe-etal-2020-translation-artifacts,riley-etal-2020-translationese-language}, and multilingual models in turn often lack the cultural and idiomatic knowledge of the target-language community \citep{hershcovich-etal-2022-cross-cultural-nlp,naous-etal-2024-cultural-bias,liu-etal-2024-cultural-reasoning,pawar-etal-2025-cultural-awareness-survey,yao-etal-2024-cultural-mt,etxaniz-etal-2024-bertaqa}.
Multilingual benchmarks and suites often contain translation-derived material, including Belebele, MMLU-ProX, XNLI, and the translated components of XTREME \citep{conneau-etal-2018-xnli,hu-etal-2020-xtreme,bandarkar-etal-2024-belebele,xuan-etal-2025-mmlu-prox}.
Native-authored benchmarks such as TyDi QA, NorEval, and INCLUDE, together with culturally annotated translated resources such as Global MMLU, provide an essential complement \citep{clark-etal-2020-tydiqa,singh-etal-2025-global-mmlu,mikhailov-etal-2025-noreval,romanou-etal-2025-include,wu-etal-2025-bitter-lesson}, while LLM-as-a-judge protocols \citep{zheng-etal-2023-mtbench,liu-etal-2023-geval,verga-etal-2024-judges-juries,gu-etal-2026-llm-judge-survey} offer a quality-sensitive complement that we exploit in Section~\ref{sec:limits}.

\section{Building MultiSynt/MT}
\label{sec:building}

\subsection{Source data: high-quality English web text}
\label{subsec:source}

We translate from the \emph{actual high-quality} split of Nemotron-CC \citep{su-etal-2025-nemotron-cc}, which contains non-synthetic Common Crawl documents receiving the highest scores from its ensemble quality classifier.
Recent reference-baseline work \citep{nezhurina-etal-2025-open-sci-ref} finds that Nemotron-CC HQ produces the strongest average downstream performance among established open English pre-training corpora across multiple model and token scales.
We draw a uniform random sample of approximately 155 million documents (on the order of 100 billion English tokens), sized so that each per-language translation yields outputs at the same order of magnitude, and apply no topical, length, or quality stratification beyond the underlying quality classifier.
Translations preserve a stable mapping to their source documents, so the resulting corpus is usable both as per-language pre-training data and as a multi-way parallel resource for cross-language and cross-system experiments (Section~\ref{subsec:overview}).

\subsection{Translation system pool and selection}
\label{subsec:systems}

We require open translation systems that combine strong target-language quality with throughput feasible at the 100B-token scale, and that together cover the languages targeted by MultiSynt/MT, including medium- and lower-resource European languages where openly available native pre-training data is scarce.
No single candidate family supplied translation models meeting all three requirements across our 36 target languages: the Tower models and the general-purpose LLM candidates evaluated in our system-selection study cover the higher-resource European subset \citep{alves-etal-2024-tower,rei-etal-2026-tower-plus,martins-etal-2025-eurollm-9b,yang-etal-2025-qwen3,gemmateam-2025-gemma-3}, while OPUS-MT and HPLT-MT extend production coverage to lower-resource languages \citep{tiedemann-etal-2024-opus-mt-democratizing,de-gibert-etal-2024-hplt,arefyev-etal-2024-hplt-models}.
We therefore use LLM-based translation and OPUS-MT/HPLT-MT jointly as the production pool.

\begin{figure}[t]
    \centering
    \includegraphics[width=\linewidth]{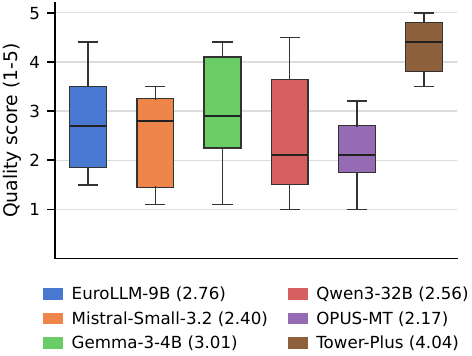}
    \caption{Mean human quality rating per candidate MT system, averaged over the seven evaluated languages on a 1--5 Likert-style scale.
    {\sc Tower+} is rated highest overall.
    See Appendix~\ref{sec:appendix-human-eval} for the protocol and Figure~\ref{fig:human-eval-per-language} for the per-language breakdown.}
    \label{fig:human-eval}
\end{figure}

To rank candidate systems, we ran a small human-judged quality study covering seven languages (Czech, French, Finnish, German, Italian, Polish, Spanish).
For each language, a single native speaker rated translations of the same 10 English source documents from each of six candidate systems on a fluency-focused 1--5 scale, blind to system identity, additionally penalizing hallucinations, repetitions, and extra or missing text.
The full protocol, KEOPS annotation interface, and per-language results are in Appendix~\ref{sec:appendix-human-eval}.
The study was designed as a small, blinded system-selection probe rather than a validation of translation quality across all 36 corpus languages, and its single-annotator design does not permit an inter-annotator agreement estimate.
{\sc Tower+} (we use the 72B model) is rated highest in every language, the general-purpose LLMs (EuroLLM-9B-Instruct, Qwen3-32B, Gemma-3-4b-it, Mistral-Small-3.2-24B-Instruct) bunch well below it, and OPUS-MT is close to the general-purpose LLMs on average (Figure~\ref{fig:human-eval}); the ranking is consistent with the public {\sc Tower} MT leaderboard and with per-language Flores-200 BLEU on the OPUS-MT dashboard.\footnote{\url{https://opus.nlpl.eu}}
Based on this ranking the general-purpose LLMs were dropped from the production pool.

We ship translations from {\sc Tower+} 9B (Gemma-based, 16 languages)\footnote{\url{https://hf.co/Unbabel/Tower-Plus-9B}} and {\sc Tower+} 72B (Qwen-based, 5 languages: German, Finnish, Italian, Spanish, Swedish).\footnote{\url{https://hf.co/Unbabel/Tower-Plus-72B}}
A reference-LLM comparison between the two on the 5-language overlap (Section~\ref{sec:effectiveness}) showed a small enough downstream gap that we treat 9B as the default LLM-based system and retain 72B on the high-resource subset where any residual gains can be isolated.
OPUS-MT supplies the production models for 31 of our 36 target languages from the Tatoeba Translation Challenge collection \citep{tiedemann-2020-tatoeba-challenge}, and HPLT-MT supplies the remaining 5 (Basque, Irish, Icelandic, Maltese, Albanian) where no comparable OPUS-MT model is available.
These NMT systems are orders of magnitude smaller than the LLM-based candidates ($\sim$75M--230M parameters) with corresponding throughput advantages, at the cost of a sentence-level constraint addressed by the pipeline of Section~\ref{subsec:pipeline}.
The released corpus therefore contains translations from {\sc Tower+} 9B (16 languages), {\sc Tower+} 72B (5 languages), and OPUS-MT/HPLT-MT (36 languages); for the languages where multiple systems are shipped, the corpus supports controlled comparison of translation choices under matched source data, which we exploit in Sections~\ref{sec:effectiveness} and~\ref{sec:limits}.

\subsection{Translation pipeline at scale}
\label{subsec:pipeline}

We run two parallel translation pipelines on the same 155 million source documents, both supporting per-shard resumption and emitting Parquet files that preserve source document identifiers and order so that the released corpus remains a synchronized multi-parallel resource.
The \emph{LLM pipeline} deploys {\sc Tower+} 9B and 72B on Leonardo HPC cluster (CINECA, NVIDIA A100 nodes) with vLLM under a Slurm-native orchestration toolkit released as open-source code,\footnote{\url{https://github.com/ellamind/inference-hive}} consuming approximately 3.1 million A100 GPU-hours in aggregate.
The \emph{NMT pipeline} decodes OPUS-MT and HPLT-MT models with Marian-NMT \citep{junczys-dowmunt-etal-2018-marian} on LUMI cluster (AMD MI250x), with sentence-splitting pre-processing and document-reassembly post-processing that preserve both per-sentence and per-document alignments; the pipeline is released as open-source code at \url{https://github.com/Helsinki-NLP/Opus-MT}.
Per-shard throughput, decoding settings, energy, and carbon-footprint estimates for both pipelines are reported in Appendix~\ref{sec:appendix-pipeline}.

\subsection{Resource overview}
\label{subsec:overview}

MultiSynt/MT contains approximately 4.8 trillion target-language tokens across 36 languages, derived from the single shared sample of 155M English documents introduced in Section~\ref{subsec:source}, and is the largest openly available pre-training corpus to date for many European low- and medium-resource languages (Figure~\ref{fig:resource-comparison}).
Counting each translation system's output separately, the released artifact totals approximately 7.1T tokens (1.7T from {\sc Tower+} 9B, 0.5T from {\sc Tower+} 72B, 4.8T from OPUS-MT/HPLT-MT, all under Gemma-3 tokenization); per-language and per-system breakdowns are in Appendix~\ref{sec:appendix-resource}.

\paragraph{Splits and schema.}
For each language we release three splits:
The \emph{parallel} split contains the documents translated by all systems released for that language, row-aligned by index so that row $i$ of any language file refers to the same source document, supporting cross-language and cross-system controlled experiments.
The \emph{additional} split contains further translated documents that are not present for every language, enabling larger language-specific volumes at the cost of cross-language alignment, and the \emph{all} split is the union of the two.
Each row contains the source \texttt{warc\_record\_id}, the translated text, its token count, and the source partition identifier.

\paragraph{Availability and license.}
MultiSynt/MT is released under a Creative Commons Zero (CC0) license applied to the corpus as a database (underlying text content remains subject to Common Crawl source-content terms), at \url{https://hf.co/datasets/MultiSynt/MT-Nemotron-CC}.
The OPUS-MT/HPLT-MT subset is additionally available on Hugging Face at \url{https://hf.co/datasets/Helsinki-NLP/nemotron-cc-translated} and as a sentence-aligned OPUS release.\footnote{\url{https://opus.nlpl.eu/synthetic/nemotron-cc-translated/v1syn}}

\section{Effectiveness for Multilingual Pre-training}
\label{sec:effectiveness}

\subsection{Reference model setup}
\label{subsec:setup}

We evaluate MultiSynt/MT as pre-training data by training reference language models on each MultiSynt/MT translation variant and on a native multilingual baseline dataset.
Across variants, we hold the model architecture, tokenizer, optimizer settings, and total token budget fixed; the variants differ in the pre-training corpus itself---specifically, the source data and the translation system (if any)---and in the set of languages each corpus covers.

Our reference architecture follows that of \citet{penedo-etal-2024-fineweb}: we use a dense Llama-like transformer with 24 layers, hidden size 2048, 32 attention heads, RoPE positional embeddings, and RMSNorm,
with 1.61B parameters in transformer layers.
We use the \texttt{google/gemma-3-27b-pt} tokenizer \citep{gemmateam-2025-gemma-3} with a vocabulary size of 262144 for its broad multilingual coverage across the languages in MultiSynt/MT.
We tie word embeddings, giving a total of 2.15B parameters.
The optimization recipe uses Adam with $\beta_1=0.9$, $\beta_2=0.95$, weight decay 0.05, gradient clipping at 1, global batch size 1024, and a linear WSD learning-rate schedule with 10\% warmup and 20\% cooldown.
Full hyperparameters are reported in Appendix~\ref{sec:appendix-hyperparams}, Table~\ref{tab:reference-hyperparams}.

Training is conducted with Megatron-LM \citep{shoeybi-etal-2019-megatron-lm} on the LUMI supercomputer, using 16 nodes with 4 AMD MI250x GPUs per node.
For every pre-training data variant, we train a separate model on a budget of 100 billion tokens, saving checkpoints every 1{,}000 steps.
Each variant is trained once; we do not report variance over training seeds, and the curves shown in Section~\ref{subsec:gains} should be read as single-seed point estimates.

We compare four pre-training data variants under this setup: HPLT 2.0 \citep{burchell-etal-2025-hplt}, a strong openly available naturally occurring multilingual web-text baseline at the relevant scale, and MultiSynt/MT translated by {\sc Tower+} 9B (16 languages), {\sc Tower+} 72B (5 languages), and OPUS-MT/HPLT-MT (36 languages).
Holding architecture, optimizer, and budget fixed lets us attribute downstream differences to the pre-training corpus.
Within the MultiSynt/MT family, this isolates the choice of translation system; between MultiSynt/MT and HPLT 2.0 the comparison additionally varies the source corpus and its quality-filtering pipeline, a confounding factor we discuss in Section~\ref{sec:discussion}.

\paragraph{Evaluation.}
Downstream evaluation uses two open frameworks, LightEval \citep{habib-etal-2023-lighteval} and LM-evaluation-harness \citep{gao-etal-2024-eval-harness}.
We evaluate each checkpoint on a broad multilingual benchmark suite: belebele \cite{bandarkar-etal-2024-belebele}, arc:challenge \cite{allenai:arc,dac2023okapi}, hellaswag \cite{zellers2019hellaswag,dac2023okapi}, goldenswag \cite{chizhov2025hellaswagvaliditycommonsensereasoning}, xstory-cloze \cite{DBLP:journals/corr/abs-2112-10668}, global-mmlu \cite{singh-etal-2025-global-mmlu}, mmlu \cite{hendrycks-etal-2021-mmlu,dac2023okapi}, exams \cite{hardalov-etal-2020-exams}, xcodah \cite{Chen2019CODAHAA,lin-etal-2021-common-sense}, xcsqa \cite{Talmor2019commonsenseqaaq,lin-etal-2021-common-sense}, MultiBlimp \cite{jumelet2026multiblimp10massivelymultilingual}, enem \cite{nunes2023evaluating}, faquad \cite{FaQuaD}, oab-exams \cite{pires2025automatic}, glianorex \cite{Griot_2025}, xnli \cite{conneau-etal-2018-xnli}, pawsx \cite{pawsx2019emnlp}, FIN-bench-v2 \cite{kytöniemi2025finbenchv2unifiedrobustbenchmark}.
We apply the same evaluation protocol across all variants.

Per-language scores are aggregated into a single mean across the languages covered by every variant under comparison; per-language and per-benchmark breakdowns are
provided as supplementary data.

\paragraph{Decontamination check.}
\label{subsec:decont}
A controlled training experiment with vs.\ without n-gram decontamination of the Nemotron-CC source rules out source-document contamination as an explanation for the Section~\ref{subsec:gains} gains (the released MultiSynt/MT corpus is itself not decontaminated due to the low risk of actual contamination; see Appendix~\ref{sec:appendix-decont}).

\subsection{Gains over native baselines}
\label{subsec:gains}

Figure~\ref{fig:gains-compute} reports aggregate benchmark performance over training tokens, comparing reference LLMs pre-trained on the HPLT 2.0 native baseline against reference LLMs pre-trained on MultiSynt/MT OPUS-MT, averaged over the five languages where both baselines have evaluation data (Danish, Dutch, Italian, Portuguese, Swedish).
Models pre-trained on MultiSynt/MT reach the HPLT 2.0 endpoint score at approximately 28B training tokens, roughly $72\%$ fewer than the native baseline; at the matched 100B-token training budget, MultiSynt/MT exceeds the HPLT 2.0 endpoint by approximately 15\% relative.

\begin{figure}[t]
    \centering
    \includegraphics[width=\linewidth]{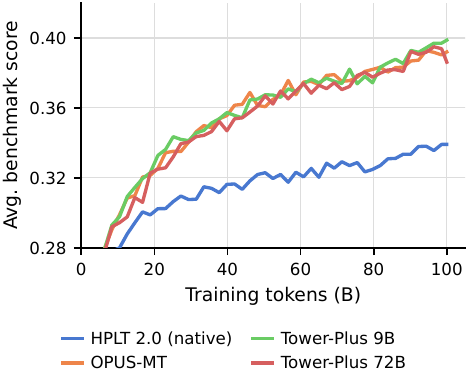}
    \caption{MultiSynt/MT variants (OPUS-MT, {\sc Tower+} 9B, {\sc Tower+} 72B) versus the HPLT 2.0 native baseline on the Swedish multilingual benchmark suite, the single language for which a reference LLM was trained on every variant.
    The three MultiSynt/MT curves sit indistinguishably above HPLT 2.0 throughout training: translated versus native dominates MT-system choice.}
    \label{fig:gains-mt-variants}
\end{figure}

Figure~\ref{fig:gains-mt-variants} compares the three MultiSynt/MT variants (OPUS-MT, {\sc Tower+} 9B, {\sc Tower+} 72B) on Swedish, the single language for which a reference LLM was trained on every variant.
The three MultiSynt/MT curves cluster tightly together throughout training and all sit well above the HPLT 2.0 baseline, indicating that on these standard multilingual benchmarks the gain over native data is largely independent of the choice of MT system within MultiSynt/MT.
The downstream gap among {\sc Tower+} 9B, {\sc Tower+} 72B, and OPUS-MT is therefore small enough that we treat the 9B variant as our default LLM-based system and retain 72B only on the high-resource subset where any residual gains can be isolated (Section~\ref{subsec:setup}).

Per-language and per-benchmark breakdowns are reported in Appendix~\ref{sec:appendix-resource}.

\section{Interpreting the Gains: Fluency, Benchmark Alignment, and Native-Authored Tasks}
\label{sec:limits}

Section~\ref{sec:effectiveness} showed that training on MultiSynt/MT yields large gains over native data on standard multilingual benchmarks for the five evaluated high- and medium-resource languages.
We now interpret those gains along three axes the standard benchmarks do not surface: an LLM-judge fluency probe that exposes MT-system differences flattened by multiple-choice scoring (Section~\ref{subsec:fluency-judge}), an embedding-space diagnostic of benchmark-corpus alignment (Section~\ref{subsec:benchmark-alignment}), and a Norwegian case study on native-authored idiomatic and culturally grounded tasks (Section~\ref{subsec:norwegian}).
We draw the methodological consequences in Section~\ref{sec:discussion}.

\subsection{What standard benchmarks miss about MT-system choice}
\label{subsec:fluency-judge}

The MT-variant indistinguishability of Figure~\ref{fig:gains-mt-variants} is surprising: the same three translation systems are clearly separated by human raters (Figure~\ref{fig:human-eval}) and by the public {\sc Tower} MT leaderboard.
A plausible explanation is that translationese artifacts in the MT output propagate to the trained LLM in a form the standard multilingual benchmarks of Section~\ref{sec:effectiveness}, predominantly multiple-choice QA and selection, cannot detect: those benchmarks reward picking the right token under strong exploitable cues, while the surface fluency of the model's own free-form generations does not enter the score.

To probe this hypothesis, we evaluate the pre-trained models with an LLM-as-a-judge protocol that scores fluency of free-form generations.
For each of German, Spanish, Finnish, Italian and Swedish we collect 100 natural-language cut-out sentences (e.g., \textit{``Das Grundgesetz der Bundesrepublik Deutschland garantiert\dots''}).
Each candidate pre-trained model produces a continuation that DeepSeek V3.1 then compares against a continuation from a monolingual HPLT 2.0 1.7B native-data baseline; we report the candidate's \emph{winrate} averaged over prompts and languages.

\begin{figure}[htb]
    \centering
    \includegraphics[width=\linewidth]{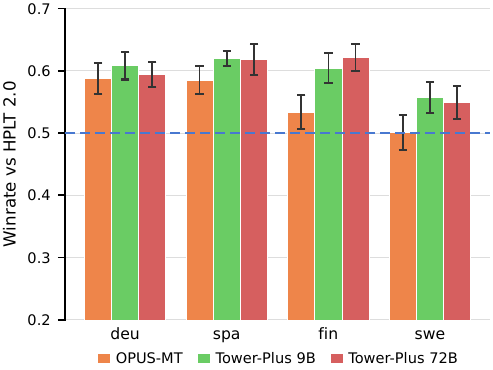}
    \caption{Fluency LLM-judge winrate of our reference 1.7B LLMs pre-trained on MultiSynt/MT against a monolingual HPLT 2.0 1.7B baseline.
    The ranking {\sc Tower+} 72B $\geq$ {\sc Tower+} 9B $>$ OPUS-MT matches the human MT evaluation (Figure~\ref{fig:human-eval}) and the public Tower MT leaderboard.
    }
    \label{fig:fluency-multisynt}
\end{figure}

As a sanity check, the off-the-shelf Qwen 2.5 series at five sizes from 0.5B to 14B parameters shows winrate rising monotonically with model size against the HPLT 2.0 baseline (Appendix~\ref{sec:appendix-fluency}, Figure~\ref{fig:fluency-qwen}), confirming the judge responds to generation quality.
The reference LLMs pre-trained on MultiSynt/MT with matched data and budget produce a clear ranking, {\sc Tower+} 72B $\geq$ {\sc Tower+} 9B $>$ OPUS-MT (Figure~\ref{fig:fluency-multisynt}), consistent with the human evaluation of the underlying MT outputs (Figure~\ref{fig:human-eval}) and the public {\sc Tower} MT leaderboard.
Under this protocol, all three MultiSynt/MT variants match or exceed the HPLT 2.0 baseline, so we detect no relative fluency deficit; the finding is that standard benchmarks are blind to translation-quality differences the trained LLMs nonetheless inherit.

\begin{figure}[htb]
    \centering
    \includegraphics[width=\linewidth]{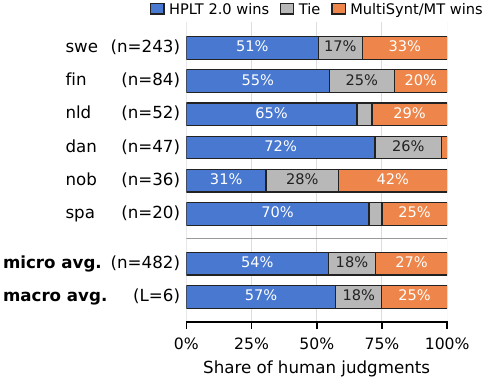}
    \caption{Human arena preferences between reference models trained on MultiSynt/MT and monolingual HPLT 2.0 baselines, pooled across the available MultiSynt/MT translation variants.
    We include languages with at least 20 judgments.
    The macro average weights each included language equally, whereas the micro average pools all 482 included judgments.}
    \label{fig:arena-stats}
\end{figure}

Additionally, we collected human fluency judgments through an arena-style blind A/B evaluation.\footnote{\url{https://github.com/OpenEuroLLM/oellm-arena}}
Raters selected the more fluent of two model continuations presented side by side or marked the comparison as a tie.
Figure~\ref{fig:arena-stats} reports comparisons between reference models trained on MultiSynt/MT and the corresponding monolingual HPLT 2.0 baselines.
Across the six languages with at least 20 judgments, HPLT 2.0 receives more wins in Swedish, Finnish, Dutch, Danish, and Spanish, whereas MultiSynt/MT receives more wins in Bokm{\aa}l.
However, sample sizes vary from 20 judgments for Spanish to 243 for Swedish, so we treat these cross-language patterns as preliminary.

\subsection{Benchmark-corpus alignment in embedding space}
\label{subsec:benchmark-alignment}
As a complementary diagnostic, we compare the embedding neighborhoods of standard benchmark items against MultiSynt/MT and HPLT 2.0 documents.
Across most benchmarks in the suite, benchmark items have more MultiSynt/MT documents among their nearest embedding neighbors, suggesting that translated high-quality English-source data provides strong coverage of many benchmark-adjacent regions; this is not a causal test of the gains, but it indicates that benchmark-corpus alignment is one reason those gains should be interpreted with care (Appendix~\ref{sec:appendix-infogap}).

\begin{figure*}[!t]
    \centering
    \includegraphics[width=0.9\linewidth]{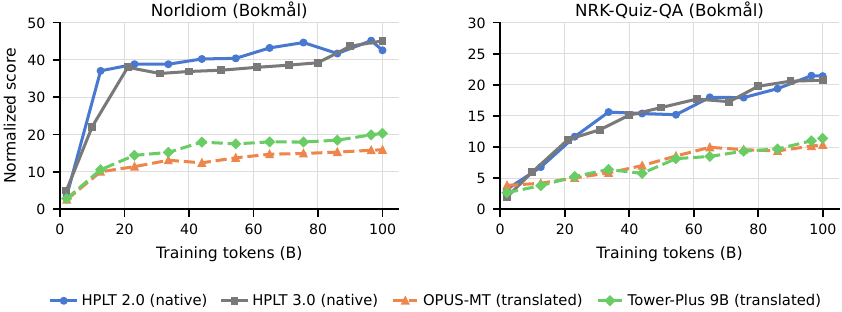}
    \caption{NorEval performance across 11 training checkpoints for reference LLMs pre-trained on native Norwegian (HPLT 2.0, HPLT 3.0) versus MultiSynt/MT translated (OPUS-MT, {\sc Tower+} 9B) data.
    Scores are normalized between the task-specific random baseline and the maximum possible score.
    Native-data models lead throughout training on Bokm{\aa}l idiom completion (left) and on questions drawn from Norwegian public-broadcaster quizzes (right).}
    \label{fig:noreval}
\end{figure*}

\subsection{Where translated data falls short: a Norwegian case study}
\label{subsec:norwegian}

Named entities, idiomatic constructions, and culturally specific knowledge in English text remain English-anchored after translation, so we expect translated data to underperform native data on target-language tasks probing local idiomaticity or culture even when the translations are fluent.
We test this using four reference LLMs that share the architecture and recipe of Section~\ref{sec:effectiveness} and differ only in Norwegian pre-training data: HPLT 2.0 and HPLT 3.0 as native data, and MultiSynt/MT translated by {\sc Tower+} 9B and OPUS-MT.
We evaluate 11 training checkpoints on NorEval \citep{mikhailov-etal-2025-noreval}, which contains 24 human-created datasets spanning Bokm{\aa}l and Nynorsk, the two official written standards of Norwegian.
Figure~\ref{fig:noreval} presents two Bokm{\aa}l tasks that directly probe the hypothesized gap: \textit{NorIdiom}, an idiom-completion task, and \textit{NRK-Quiz-QA}, a multiple-choice task based on quizzes from the Norwegian public broadcaster and targeting Norwegian-specific and world knowledge.
Appendix~\ref{sec:appendix-noreval} gives the complete evaluation parameterization, the corresponding Nynorsk results, and an aggregate over all 33 task configurations in the updated evaluation export.

On \textit{NorIdiom} (Figure~\ref{fig:noreval}, left), both native-data models outperform both translated-data models with a stable gap throughout training, consistent with idiomatic constructions being a long-tail phenomenon of the target language that the translated corpus does not generate at the rate native text does.
On \textit{NRK-Quiz-QA} (Figure~\ref{fig:noreval}, right), the native-data models also lead throughout training, consistent with the English source corpus providing limited exposure to Norwegian cultural knowledge.
The Nynorsk variants reproduce both patterns, and the aggregate across the broader NorEval suite likewise favors native data (Appendix~\ref{sec:appendix-noreval}).
This result does not negate the multilingual benchmark gains of Section~\ref{sec:effectiveness}: the Norwegian adaptations of CommonsenseQA and OpenBookQA favor translated-data models at later checkpoints, echoing those standard benchmarks and indicating that native and translated pre-training data contribute different strengths (Appendix~\ref{sec:appendix-noreval}).

\section{Discussion and Recommendations}
\label{sec:discussion}

\paragraph{Treat translated data as a complement to native data, not a replacement.}
For high-resource languages, where native corpora are already abundant, MultiSynt/MT is most useful as a domain-coverage extension.
For medium- and lower-resource languages, MultiSynt/MT substantially expands the available corpus volume, but its downstream effectiveness remains unverified for the lowest-resource languages in the release.
The tested Norwegian setting motivates pairing translated data with available native corpora to preserve idiomatic and cultural anchoring, while the localized standard tasks suggest that translated data contributes complementary transferable knowledge (Section~\ref{subsec:norwegian}).

\paragraph{The headline gain may reflect source quality alone, not translation as a beneficial operation.}
The Section~\ref{subsec:gains} comparison varies both the source corpus (Nemotron-CC HQ vs.\ native HPLT 2.0 web text) and the translation step, so the gain cannot be attributed to translation \emph{per se}: it is consistent with the net effect arising entirely from the higher quality of the English source, with translation contributing nothing of its own.
Disentangling the two would require a controlled source-corpus comparison that we leave to future work.
The practical counterfactual is nonetheless not arbitrary translated versus arbitrary native text, but translated high-quality source text at 100B-token scale versus the much smaller native pool that would survive equivalently stringent quality filtering, which for medium- and lower-resource languages would shrink the already-scarce native data to a small fraction of itself.

\paragraph{Pair translation-derived benchmarks with native-authored ones, and complement multiple-choice with quality-sensitive measures.}
A large fraction of widely used multilingual benchmarks, including Belebele \citep{bandarkar-etal-2024-belebele}, m-ArenaHard \citep{dang-etal-2024-aya-expanse}, MMMLU \citep{openai-2024-mmmlu} and MT-Bench-X \citep{weber-etal-2024-mt-bench-x}, are themselves produced by translating English sources, and our embedding-space diagnostic shows that they tend to sit in neighborhoods with more MultiSynt/MT than HPLT 2.0 documents (Section~\ref{subsec:benchmark-alignment}).
A model trained on translated data shares the same source-language frame as these benchmarks and is, in part, being evaluated on how well it has absorbed that frame; standard multiple-choice scoring is additionally blind to fluency-quality differences that the resulting LLMs nevertheless inherit (Section~\ref{subsec:fluency-judge}).
Native-authored benchmarks such as NorEval \citep{mikhailov-etal-2025-noreval} expose gaps that translation-derived benchmarks fail to surface (Section~\ref{subsec:norwegian}); free-form quality-sensitive measures such as the LLM-judge protocol of Section~\ref{subsec:fluency-judge} expose differences that discriminative ones flatten.

\section{Conclusion}
\label{sec:conclusion}

MultiSynt/MT provides approximately 4.8T target-language tokens across 36 languages, the largest openly available pre-training corpus to date for many European low- and medium-resource languages.
Across the five evaluated high- and medium-resource languages at our single reference-model scale, models trained on it match a native-data baseline (HPLT 2.0) using roughly 72\% fewer pre-training tokens and exceed it by approximately 15\% relative at a matched 100B-token budget.
The translated corpus nonetheless falls short on idiomatic and culturally grounded Norwegian tasks and is best understood as a complement to native data; we release it under CC0 with row-aligned translations from multiple MT systems to support controlled follow-up on translation-system choice, scaling, native-benchmark coverage, and source-quality versus translation attribution.

\section*{Limitations}

\paragraph{Corpus and translation biases.}
MultiSynt/MT inherits the topical, stylistic, and English-centric biases of its Nemotron-CC HQ source: a quality-filtered Common Crawl pool that favours factual prose and formal registers over oral, vernacular, dialectal, and culturally local text, and that anchors named entities, idioms, and culturally specific knowledge to an English reference frame.
The chosen MT systems further introduce systematic translationese artifacts (reduced lexical variety, calque-driven syntax, lower idiomatic density) that the standard multilingual benchmarks of Section~\ref{sec:effectiveness} flatten but that surface under fluency-sensitive evaluation (Section~\ref{subsec:fluency-judge}) and on natively authored idiomatic and culturally grounded tasks (Section~\ref{subsec:norwegian}).
The headline comparison against HPLT 2.0 (Section~\ref{subsec:gains}) additionally varies the source distribution and its quality-filtering pipeline, so the gain cannot be attributed to translation alone; an isolating comparison against an equivalently stringently filtered native baseline is left to future work.

\paragraph{Evaluation coverage, scale, and feedback risks.}
The headline downstream comparison covers five high- and medium-resource European languages and does not directly verify effectiveness for the lowest-resource languages that receive the largest corpus-volume increases.
Our experiments are conducted at a single reference-model scale, with 1.61B non-embedding parameters and 2.15B parameters in total, and use single-seed pre-training runs (Section~\ref{subsec:setup}), so scaling trends to larger models and seed-level variance are not directly verified.
The embedding-space alignment diagnostic (Section~\ref{subsec:benchmark-alignment}) is descriptive rather than causal, and the fluency LLM-judge protocol (Section~\ref{subsec:fluency-judge}) is conditional on its judge and prompt set.
The human MT evaluation (Section~\ref{subsec:systems}) is a system-selection probe covering seven languages, ten documents, and one annotator per language, and therefore provides neither corpus-wide validation nor inter-annotator agreement estimates.
The arena evaluation is likewise preliminary because judgments are unevenly distributed across languages and concentrated in Swedish.
Finally, releasing MultiSynt/MT openly creates a feedback path into future open MT systems trained on web crawls that will then ingest it, with the attendant long-tail erosion risks of recursive synthetic-data training \citep{shumailov-etal-2024-model-collapse}; we mark the corpus as synthetic in the released metadata and recommend treating it as a complement to native data rather than a replacement (Section~\ref{sec:discussion}).

\section*{Acknowledgments}

We acknowledge EuroHPC JU for awarding the project ID EHPC-AIF-2025LS01-028 access to the EuroHPC supercomputer LEONARDO hosted by CINECA (Italy) and the LEONARDO consortium.
We acknowledge EuroHPC JU for awarding the project ID EHPC-AIF-2025LS16-024 access to the EuroHPC supercomputer MareNostrum 5 hosted by the Barcelona Supercomputing Center (BSC).
We acknowledge EuroHPC JU for awarding the project ID HPC-REG-2024R02-167 access to the EuroHPC supercomputer LUMI hosted by CSC (Finland) and the LUMI consortium.
Part of the computations were performed on resources provided by Sigma2 - the National Infrastructure for High-Performance Computing and Data Storage in Norway.
This research was supported by the OpenEuroLLM project, co-funded by the Digital Europe Programme under GA no. 101195233, and partially by DECOLLAGE (ERC-2022-CoG 101088763).
This project is supported by the German Federal Ministry for Economic Affairs and Energy (BMWE) through EU-SAI/SOOFI: Sovereign Open Source Foundation Models for European Intelligence (grant number 13IPC040J).
This research was supported by the Research Council of Finland through the Human Diversity consortium, under the Profi7 program (grant no. 352727) and the Centre of Excellence (grant no. 374221).
This research was supported by the European Union's Horizon Europe research and innovation programme under the Marie Skłodowska-Curie grant agreement No. 101126636.
\bibliography{reference.bib}

\appendix

\section{Translation Pipeline Details}
\label{sec:appendix-pipeline}

\paragraph{Orchestration toolkit (LLM pipeline).}
For the LLM-based translation runs of Section~\ref{subsec:pipeline}, we use an open-source Slurm-native orchestration toolkit\footnote{\url{https://github.com/ellamind/inference-hive}} that scales inference workloads across hundreds to thousands of GPUs while maintaining linear throughput.
The toolkit deploys multiple distributed inference servers in parallel: a single YAML configuration file specifies the cluster, inference-server, and data settings, and the toolkit generates the required Slurm job scripts, manages resource allocation, and coordinates distributed execution.
It provides error handling with automatic retries and resumption, progress and throughput monitoring, built-in support for large-scale loading of Parquet input files, and a flexible inference server backend.
We ran all inference with vLLM \citep{kwon2023vllm}.

\paragraph{LLM runs: per-model throughput.}
Across batched production runs (4 GPUs per inference server), {\sc Tower+} 9B sustained approximately 1{,}000 output tokens per second per GPU, and {\sc Tower+} 72B approximately 400 output tokens per second per GPU.
The 9B variant therefore produces output tokens approximately $2.6\times$ faster than the 72B variant per GPU-second; the per-request advantage is wider (approximately $3.2\times$) because the 72B model supports a longer maximum sequence length, allowing us to feed longer source documents per request and yielding roughly 26\% longer translated outputs on average.

\paragraph{LLM runs: aggregate compute, energy and carbon footprint.}
The LLM-based translation runs were carried out on the Leonardo (CINECA) supercomputer.
Aggregate GPU consumption is approximately 3.1 million A100 GPU-hours.
We estimate the corresponding energy and carbon footprint following the methodology used in CINECA project reporting: each A100 on Leonardo draws approximately 440~W under load, giving 1{,}379{,}638~kWh of consumed energy; applying the 2025 Italian electricity-grid carbon intensity of 358~g~CO$_2$-eq/kWh yields approximately 493.9~t~CO$_2$-eq.
For reference, this energy budget corresponds to roughly 16\% of the annual generation of an average wind turbine (3~MW nameplate capacity, 2{,}800 equivalent full-load hours per year, $\approx$8.4~GWh/year).
These figures cover only the LLM-based translation pipeline; the corresponding energy and carbon estimates for the OPUS-MT/HPLT-MT runs on LUMI (AMD MI250x) are reported alongside the NMT shard statistics below.

\paragraph{NMT models: training data and inference setup.}
The OPUS-MT and HPLT-MT models we use are decoded with Marian-NMT \citep{junczys-dowmunt-etal-2018-marian} at beam size~4 on single-node jobs allocating two AMD MI250x GPUs each, on the LUMI supercomputer.
The selected releases are based on OPUS data \citep{tiedemann-2009-opus,tiedemann-2012-opus} and the Tatoeba Translation Challenge model collection \citep{tiedemann-2020-tatoeba-challenge}; individual releases may additionally incorporate back-translated Wikimedia content, as documented in the model inventory distributed with the dataset.
None of the NMT models are fine-tuned for specific tasks; we use them off-the-shelf without further data normalisation or filtering.
The final list of selected NMT models is published together with the dataset card.
The sentence-splitter pre-processing step greedily merges segments shorter than 256 characters and splits sentences longer than 1024 characters before feeding them to the decoder; this length normalisation matters because NMT batching benefits sharply from uniform input length.

\paragraph{NMT runs: sharding, runtimes, energy and per-GPU power.}
The corpus was split into 100 shards per language of approximately 30 million sentence-aligned lines (roughly 700 million space-separated tokens each).
Per-shard runtime averaged about 23 hours for transformer-base models ($\sim$75M parameters) and about 30 hours for transformer-big models ($\sim$230M parameters), with sustained GPU utilisation between 85\% and 95\%.
Total energy consumption for one shard translated with transformer-base models is roughly 12~kWh, at an average of 260--270~W per GPU.
For transformer-big models, total energy consumption is roughly double at 24--26~kWh, at an average of 360--370~W per GPU.
The pattern suggests that further batching and inference optimisation could improve GPU utilisation for the smaller models and yield additional speed-ups.

\section{Reference Model Hyperparameters}
\label{sec:appendix-hyperparams}

\paragraph{Compute setup.}
All reference-model training and evaluation runs were carried out on the LUMI supercomputer (16 nodes of 4 AMD MI250x GPUs per node), with one exception: the Nemotron-CC decontamination control experiment of Appendix~\ref{sec:appendix-decont} was run on Leonardo (CINECA) with NVIDIA A100 nodes.

\begin{table}[ht]
    \centering
    \small
    \begin{tabular}{lr}
    \toprule
    \textbf{Hyperparameter} & \textbf{Value} \\
    \midrule
    Number of layers & 24 \\
    Hidden size & 2048 \\
    Attention heads & 32 \\
    Initialization std & 0.02 \\
    Sequence length & 2048 \\
    Rotary base & 10000 \\
    Optimizer & Adam \\
    Adam $\beta_1$, $\beta_2$ & 0.9, 0.95 \\
    Adam $\epsilon$ & $1.0\times 10^{-8}$ \\
    Learning rate & $3.0\times 10^{-4}$ \\
    Minimum learning rate & 0 \\
    Learning-rate decay & linear WSD \\
    Warmup fraction & 0.1 \\
    Cooldown fraction & 0.2 \\
    Weight decay & 0.05 \\
    Gradient clipping & 1 \\
    Global batch size & 1024 \\
    Micro batch size & 4 \\
    Position embeddings & RoPE \\
    Normalization & RMSNorm \\
    Precision & bfloat16 \\
    Data parallel size & 128 \\
    \bottomrule
    \end{tabular}
    \caption{Reference-model hyperparameters used in the main pre-training runs.}
    \label{tab:reference-hyperparams}
\end{table}

\section{Decontamination: Methodology and Full Tables}
\label{sec:appendix-decont}

\paragraph{Motivation and prior work.}
Large web-derived corpora can overlap with evaluation benchmarks through exact duplicates, near-duplicates, paraphrases, or translated variants.
The GPT-3 family established n-gram-based decontamination as a standard pre-training practice \citep{brown-etal-2020-few-shot}, and deduplication has been shown to reduce memorization and improve downstream model quality \citep{lee-etal-2022-deduplication}.
Recent work emphasizes that contamination is difficult to define uniformly and that its measured impact depends on the benchmark, model, and detection setup \citep{singh-etal-2024-contamination}.
For multilingual settings the problem is sharper: contamination can cross language boundaries, and translated benchmark items can evade simple text-overlap detection while still inflating evaluation scores \citep{yao-etal-2024-crosslingual-contamination}.
We therefore decontaminate the English source documents before translation, report overlap statistics per benchmark, and run a controlled training comparison to verify that removing flagged documents has no measurable downstream effect.

\paragraph{Released corpus.}
The released MultiSynt/MT corpus itself is not decontaminated; the decontamination described here is applied to the English source documents used as input to the controlled training comparison below.

\paragraph{Pipeline stages.}
We use the NeMo Curator decontamination pipeline, which constructs n-gram representations of downstream task datasets and removes any matching content from the training corpus in three stages.
\textbf{Indexing:} each benchmark dataset is normalised and tokenised, and n-grams of size 8--13 are extracted from the test split (or the validation split, where no test split is available), following the text formatting in lm-evaluation-harness, to build a compact n-gram index.
\textbf{Matching:} each source document is processed through the same normalisation/tokenisation/n-gram-extraction pipeline and its n-grams are compared against the benchmark index to identify overlaps.
\textbf{Removal:} source documents whose matched n-grams occur fewer than 10 times in the corpus are treated as contamination signals and removed in full.

The indexed evaluation suite
(see Table~\ref{tab:decont-benchmarks-merged} for details)
spans code, common sense reasoning, instruction following, language understanding, linguistic competence, mathematics, reasoning, translation, and world knowledge.

\paragraph{Match counts and rates.}
After applying the pipeline to the source Nemotron-CC documents used as input to translation, a total of 34{,}969 documents per language (0.0002\%) are flagged as contaminated.
Per-benchmark match counts and contamination rates
are reported in
Table~\ref{tab:decont-benchmarks-merged}.
Contamination is unevenly distributed: the largest absolute contribution comes from BoolQ, followed at lower rates by HellaSwag, MMLU, and MTBench.
Even where the absolute number of matches is large, the corresponding percentages remain very low, indicating that contamination is sparse relative to corpus size.

\begin{figure*}[t]
    \centering
    \includegraphics[width=.65\linewidth]{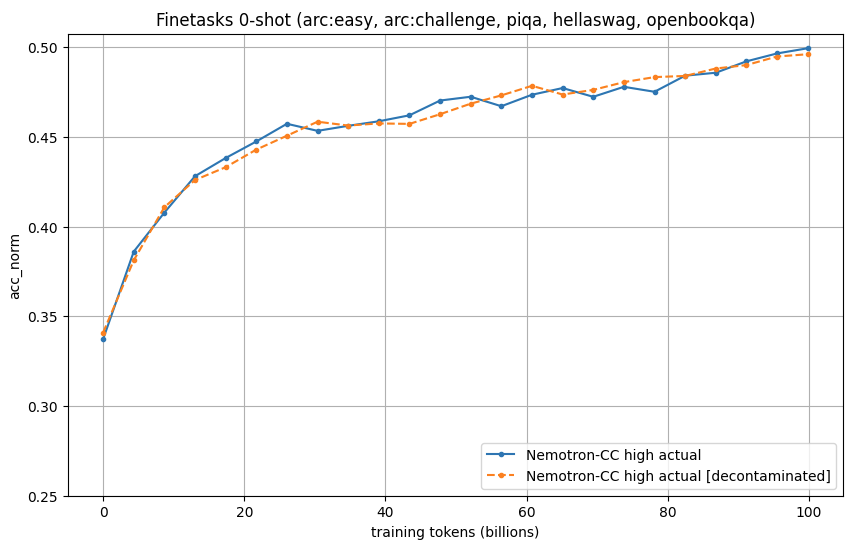}
    \caption{Controlled training experiment on the Nemotron-CC source data, with and without decontamination.
    A 0.6B-parameter reference model is trained for 100B tokens and evaluated 0-shot on the Finetasks suite (ARC-easy, ARC-challenge, PIQA, HellaSwag, OpenBookQA).
    Curves track each other throughout training, indicating no measurable downstream effect from decontamination.}
    \label{fig:decont-nemotron-results}
\end{figure*}

\paragraph{Controlled training experiment.}
To verify that the residual contamination has no measurable downstream effect, we train a 0.6B-parameter reference model on the chosen Nemotron-CC source slice with and without decontamination, under identical hyperparameters and a 100B-token budget, and evaluate 0-shot on the Finetasks suite (ARC-easy, ARC-challenge, PIQA, HellaSwag, OpenBookQA).
The two learning curves track each other throughout training (Figure~\ref{fig:decont-nemotron-results}), with no consistent advantage for either setting.

\begin{table*}[t]
    \centering
    \small
    \begin{tabular}{|l|l|r|r|r|r|}
    \hline
    \textbf{Task} & \textbf{Category} & \textbf{Passages} & \textbf{Size (MB)} & \textbf{Matches} & \textbf{Contamination\%} \\
    \hline
Humaneval & Code & 328 & 10.6419 & 20 & 1.3e-05\\
MBPP & Code & 878 & 4.8529 & 32 & 2.1e-05\\
ARC Challenge & Common sense & 3,516 & 1.1415 & 109 & 7.0e-05\\
ARC Easy & Common sense & 7,128 & 1.9972 & 157 & 1.0e-04\\
BoolQ & Common sense & 3,270 & 2.0204 & 23275 & 1.5e-02\\
CommonsenseQA & Common sense & 1,221 & 0.2635 & 0 & 0 \\
COPA & Common sense & 1,100 & 0.1422 & 0 & 0 \\
HellaSwag & Common sense & 10,042 & 10.6699 & 3034 & 2.0e-03\\
Lambada & Common sense & 5,153 & 1.6316 & 28 & 1.8e-05 \\
OpenBookQA & Common sense & 2,000 & 0.3851 & 1 & 6.5e-07\\
PIQA & Common sense & 5,512 & 1.4174 & 2 & 1.3e-06 \\
XStoryCloze & Common sense & 7,555 & 2.5407 & 0 & 0 \\
XWinograd & Common sense & 2,671 & 0.3142 & 31 & 2.0e-05 \\
AlpacaEval & Instruction & 1,593 & 0.1720 & 170 & 1.1e-04 \\
do\_not\_answer & Instruction & 939 & 3.8161 & 1 & 6.5e-07 \\
IFEval & Instruction & 541 & 0.2743 & 59 & 3.8e-05\\
m-ArenaHard & Instruction & 6,000 & 3.3170 & 14 & 9.0e-06 \\
MT-Bench-X & Instruction & 400 & 0.2720 & 44 & 2.8e-05\\
MTBench & Instruction & 1,388 & 1.8000 & 3127 & 2.0e-03 \\
Multi-IFEval & Instruction & 22,505 & 34.3774 & 2 & 1.3e-06 \\
Toxigen & Instruction & 940 & 0.3500 & 16 & 1.0e-05 \\
PAWS-X & Lang.\ underst. & 14,000 & 3.4758 & 136 & 8.8e-05\\
MultiBlimp & Linguistic & 60,693 & 92.1014 & 135 & 8.7e-05\\
AIME 25 & Math & 30 & 0.0118 & 0 & 0\\
GSM8K & Math & 1,319 & 0.6815 & 11 & 7.1e-06\\
Hendrycks\_MATH & Math & 5,003 & 2.0000 & 1065 & 6.9e-04\\
PolyMath & Math & 3,000 & 0.6709 & 64 & 4.1e-05\\
XNLI & Reasoning & 27,449 & 5.2555 & 31 & 2.0e-05 \\
FLORES-200 & Translation & 37,444 & 10.3435 & 51 & 3.3e-05\\
Belebele & World Knowl. & 29,700 & 27.5175 & 513 & 3.3e-04\\
GPQA & World Knowl. & 546 & 4.0179 & 0 & 0\\
MMLU & World Knowl. & 14,042 & 6.6514 & 2841 & 1.8e-03\\
\hline
    \end{tabular}
\caption{Benchmark evaluation suite grouped by capability category, showing approximate passage count and size of the splits used for decontamination, total number of contaminated documents containing at least one benchmark n-gram that failed the frequency threshold and ratio of contaminated documents to the full source corpus per benchmark.}
\label{tab:decont-benchmarks-merged}
\end{table*}

\section{Per-Language Resource Statistics}
\label{sec:appendix-resource}

Table~\ref{tab:per-language-tokens} reports per-language Gemma-3 token counts for each MultiSynt/MT translation system and for the HPLT 3.0 native baseline, covering all 36 languages of the released corpus.\footnote{Albanian is listed as \texttt{sqi} in MultiSynt/MT (the ISO 639-3 macrolanguage code, matching OPUS-MT/HPLT-MT's output language label) and as \texttt{als\_Latn} in HPLT 3.0 (Tosk Albanian, the basis of Standard Albanian and the FLORES/NLLB convention for the language).
The two refer to the same standard written language in practice.}
Figure~\ref{fig:resource-comparison} in Section~\ref{subsec:overview} visualizes a 24-language representative subset drawn from this table; the table itself gives the full picture, including languages where Tower+ 9B or 72B was not run and languages without an HPLT 3.0 baseline.
For consistency with Figure~\ref{fig:resource-comparison}, the MultiSynt/MT-to-HPLT 3.0 ratio is computed against the per-language maximum across MultiSynt/MT systems (effectively OPUS-MT/HPLT-MT, whose output exceeds Tower+ on every language).
Sums of the per-system columns recover the corpus totals reported in Section~\ref{subsec:overview}: 1.7T from Tower+ 9B (16 languages), 0.5T from Tower+ 72B (5 languages), 4.8T from OPUS-MT/HPLT-MT (36 languages), for approximately 7.1T tokens across all system outputs and approximately 4.8T unique target-language tokens.

\begin{table*}[t]
    \centering
    \small
    \begin{tabular}{llrrrrr}
    \toprule
    \textbf{Code} & \textbf{Language} & \textbf{Tower+ 9B} & \textbf{Tower+ 72B} & \textbf{OPUS-MT} & \textbf{HPLT 3.0} & \textbf{MS/HPLT} \\
    \midrule
    mlt & Maltese            & --    & --    & 178.4 & 1.0   & 182$\times$ \\
    gle & Irish              & --    & --    & 167.0 & 1.1   & 153$\times$ \\
    nno & Norwegian Nynorsk  & 107.6 & --    & 124.8 & 1.6   & 79$\times$ \\
    glg & Galician           & --    & --    & 112.0 & 3.1   & 36$\times$ \\
    eus & Basque             & --    & --    & 127.9 & 3.2   & 40$\times$ \\
    mkd & Macedonian         & --    & --    & 154.6 & 5.9   & 26$\times$ \\
    isl & Icelandic          & 136.8 & --    & 152.1 & 6.1   & 25$\times$ \\
    kat & Georgian           & --    & --    & 172.0 & 7.5   & 23$\times$ \\
    sqi & Albanian           & --    & --    & 161.8 & 10.1  & 16$\times$ \\
    srp & Serbian            & --    & --    & 148.1 & 10.4  & 14$\times$ \\
    lav & Latvian            & --    & --    & 154.2 & 17.2  & 8.9$\times$ \\
    est & Estonian           & --    & --    & 132.7 & 20.6  & 6.4$\times$ \\
    slv & Slovenian          & --    & --    & 133.0 & 20.9  & 6.4$\times$ \\
    cat & Catalan            & --    & --    & 131.3 & 22.5  & 5.8$\times$ \\
    lit & Lithuanian         & --    & --    & 145.1 & 28.8  & 5.0$\times$ \\
    bos & Bosnian            & --    & --    & 113.8 & 32.0  & 3.6$\times$ \\
    hrv & Croatian           & --    & --    & 125.5 & 35.1  & 3.6$\times$ \\
    slk & Slovak             & --    & --    & 135.9 & 40.2  & 3.4$\times$ \\
    bul & Bulgarian          & --    & --    & 139.7 & 49.0  & 2.9$\times$ \\
    nob & Norwegian Bokm\aa{}l & 103.6 & --  & 119.7 & 51.2  & 2.3$\times$ \\
    dan & Danish             & 106.1 & --    & 122.0 & 62.7  & 1.9$\times$ \\
    fin & Finnish            & 120.0 & 121.4 & 140.1 & 73.9  & 1.9$\times$ \\
    ukr & Ukrainian          & 116.0 & --    & 135.6 & 81.2  & 1.7$\times$ \\
    hun & Hungarian          & 122.3 & --    & 141.0 & 102.3 & 1.4$\times$ \\
    ron & Romanian           & 112.8 & --    & 131.8 & 102.5 & 1.3$\times$ \\
    swe & Swedish            & 103.3 & 109.0 & 118.6 & 111.8 & 1.06$\times$ \\
    ell & Greek              & --    & --    & 174.2 & 115.6 & 1.5$\times$ \\
    ces & Czech              & --    & --    & 128.4 & 126.2 & 1.02$\times$ \\
    tur & Turkish            & --    & --    & 115.9 & 150.0 & 0.77$\times$ \\
    nld & Dutch              & 102.5 & --    & 118.1 & 173.4 & 0.68$\times$ \\
    pol & Polish             & 110.2 & --    & 126.8 & 270.1 & 0.47$\times$ \\
    por & Portuguese         &  93.5 & --    & 107.5 & 318.9 & 0.34$\times$ \\
    ita & Italian            &  97.2 & 102.3 & 111.9 & 335.5 & 0.33$\times$ \\
    fra & French             & 102.2 & --    & 115.0 & 585.0 & 0.20$\times$ \\
    deu & German             &  98.7 & 100.2 & 112.1 & 609.3 & 0.18$\times$ \\
    spa & Spanish            &  91.3 &  97.8 & 104.3 & 659.0 & 0.16$\times$ \\
    \midrule
    \multicolumn{2}{l}{\textbf{Total}} & 1724.0 & 530.7 & 4833.1 & 4245.0 & 1.14$\times$ \\
    \bottomrule
    \end{tabular}
    \caption{Per-language Gemma-3 token counts (billions) for each MultiSynt/MT translation system and the HPLT 3.0 native baseline, for the 36 languages of MultiSynt/MT.
    Rows are sorted by HPLT 3.0 token count in ascending order.
    The MS/HPLT column gives the ratio of the per-language MultiSynt/MT maximum (always OPUS-MT/HPLT-MT, whose output exceeds Tower+ on every language) to HPLT 3.0.
    The Total row sums each column; the Total MS/HPLT ratio is the sum of MultiSynt/MT maxima divided by the sum of HPLT 3.0 tokens.
    Tower+ 9B was run on 16 languages, Tower+ 72B on a five-language high-resource subset, and OPUS-MT/HPLT-MT on all 36 languages.}
    \label{tab:per-language-tokens}
\end{table*}

\section{Human MT evaluation: per-language breakdown}
\label{sec:appendix-human-eval}

\begin{figure*}[t]
    \centering
    \includegraphics[width=\linewidth]{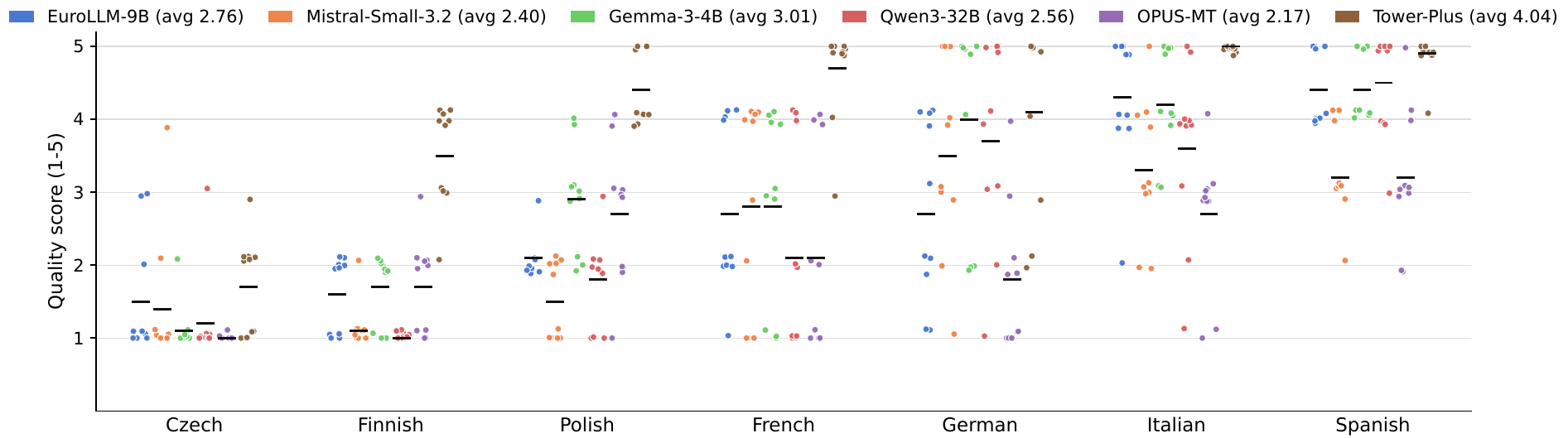}
    \caption{Per-language breakdown of the human evaluation of MT quality (cf.\ Figure~\ref{fig:human-eval}).
    Languages are arranged along the x-axis by mean quality across all six systems (hardest on the left, easiest on the right); the per-system overall mean across the seven languages is shown next to each legend entry.
    {\sc Tower+} is rated highest in every language; per-language deviations among the other systems are detailed in the bars.}
    \label{fig:human-eval-per-language}
\end{figure*}

Figure~\ref{fig:human-eval-per-language} shows the per-language breakdown of the human evaluation summarized in Section~\ref{subsec:systems} and Figure~\ref{fig:human-eval}.

\paragraph{Protocol.}
For each language, a single native speaker of the target language rated translations of the same 10 English source documents from each of the six candidate systems, blind to which system produced each translation; because the source documents are held constant across systems within a language, the within-language comparison among systems is paired.
Annotators were instructed to focus primarily on target-language fluency and to additionally penalize specific model failures (hallucinations, repetitions, extra text not present in the source, and missing text); machine-translation outputs that were truncated due to inference length constraints were flagged in the guidelines and were exempted from penalty.
Annotators recorded each judgment on a continuous slider mapped to a 0--100 scale, with the endpoints anchored as \emph{very poor translation} (0) and \emph{excellent translation} (100); the recorded scores are aggregated into five 20-point bins (0--20, 21--40, 41--60, 61--80, 81--100) and reported on a 1--5 Likert-style scale (1 = bad, 5 = excellent).

\paragraph{Results.}
{\sc Tower+} is rated highest in each of the seven evaluated languages; the relative ordering of the other systems varies by language.
Czech and Finnish, the two morphologically richest languages in the evaluated set, receive the lowest scores across all systems, including {\sc Tower+}; on the higher-resource Romance languages (French, Italian, Spanish) the gap between {\sc Tower+} and the general-purpose LLMs narrows substantially, while OPUS-MT remains a clear step below the LLM-based systems.

\paragraph{Comparison to WMT25.}
The final WMT25 human evaluation \citep{kocmi-etal-2025-wmt25-findings} places {\sc Tower+} behind the strongest proprietary systems on several evaluated high-resource directions, with its rank varying substantially by language pair.
That evaluation, however, operates on paragraph-sized segments of approximately 100 words drawn from news, social media, speech transcripts, and literary text, rather than on the full web documents that constitute the Nemotron-CC HQ source distribution; several of the higher-ranked systems are also closed-weights or otherwise infeasible to deploy at the 100B-token source scale on the open compute available for this work.

\section{NorEval Configuration and Additional Results}
\label{sec:appendix-noreval}

\paragraph{Evaluation configuration.}
All NorEval results use zero-shot evaluation across 11 checkpoints per model.
For each task and model, we select the maximum score across the available human-written prompts and across supported task formulations, including CF, MCF, and hybrid formulations where applicable.
We use normalized accuracy for multiple choice tasks and the task's primary metric for generative tasks.
Following the NorEval aggregation protocol \citep{mikhailov-etal-2025-noreval}, we rescale each score between its task-specific random baseline and maximum possible value.
The full-suite result averages these normalized scores over all 33 task configurations in the updated evaluation export.

\begin{figure*}[t]
    \centering
    \includegraphics[width=\linewidth]{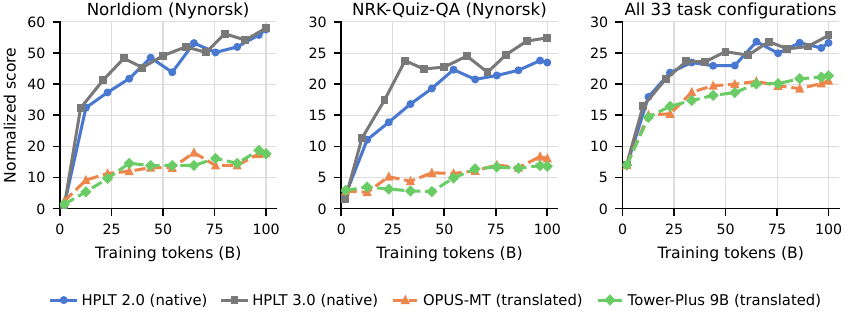}
    \caption{Additional NorEval results under the configuration described in Appendix~\ref{sec:appendix-noreval}.
    The Nynorsk variants of NorIdiom (left) and NRK-Quiz-QA (center) reproduce the native-data advantage observed for Bokm{\aa}l in Figure~\ref{fig:noreval}.
    The aggregate over all 33 evaluated task configurations (right) likewise favors native Norwegian pre-training data.}
    \label{fig:noreval-nynorsk-aggregate}
\end{figure*}

\paragraph{Nynorsk and aggregate results.}
Figure~\ref{fig:noreval-nynorsk-aggregate} shows that the native-data advantage on idioms and locally grounded quiz questions extends from Bokm{\aa}l to Nynorsk.
The aggregate over both written standards and all 33 task configurations also places both native-data models above both translated-data models at the final checkpoint.

\begin{figure*}[t]
    \centering
    \includegraphics[width=0.9\linewidth]{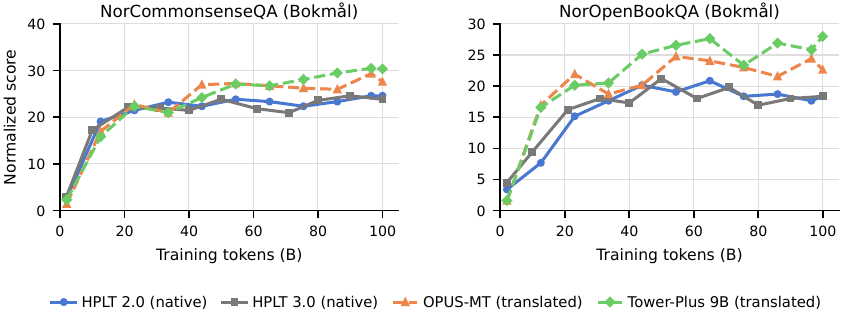}
    \caption{NorEval bridge tasks based on localized standard English-origin benchmarks.
    On Bokm{\aa}l NorCommonsenseQA (left) and NorOpenBookQA (right), the translated-data models lead at later checkpoints, echoing the multilingual benchmark results in Section~\ref{sec:effectiveness}.}
    \label{fig:noreval-bridge}
\end{figure*}

\paragraph{Bridge to translated standard benchmarks.}
NorCommonsenseQA and NorOpenBookQA are manually localized Norwegian variants of standard English-origin tasks and are therefore closer in construction to the multilingual benchmarks in Section~\ref{sec:effectiveness} than NorIdiom and NRK-Quiz-QA.
Figure~\ref{fig:noreval-bridge} shows that both translated-data models lead the native-data models at later checkpoints on these tasks.
These results reproduce the direction of the multilingual benchmark comparison within the Norwegian study and support a complementary interpretation: translated high-quality English data transfers well to localized standard tasks, whereas native data better supports language-specific idioms and locally grounded knowledge.

\section{Fluency LLM-judge: sanity check}
\label{sec:appendix-fluency}

\begin{figure}[ht]
    \centering
    \includegraphics[width=\linewidth]{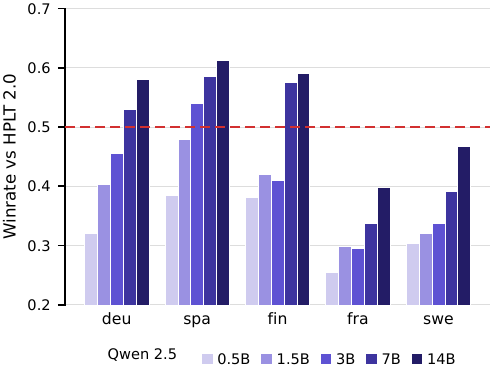}
    \caption{Sanity check: fluency LLM-judge winrate of the Qwen 2.5 model series (0.5B--14B parameters) against monolingual HPLT 2.0 1.7B baselines trained on native data.
    Winrate rises monotonically with model size across all evaluated languages, confirming that the judge tracks generation quality.
    The dashed line at 0.5 marks parity with the native baseline.}
    \label{fig:fluency-qwen}
\end{figure}

Figure~\ref{fig:fluency-qwen} reports the sanity check referenced in Section~\ref{subsec:fluency-judge}: the off-the-shelf Qwen 2.5 series at five sizes from 0.5B to 14B parameters, evaluated by the same LLM-judge protocol against the monolingual HPLT 2.0 1.7B baseline.
Winrate against the baseline rises monotonically with Qwen model size across all evaluated languages, confirming that the judge tracks generation quality and that the MT-variant ranking reported in Figure~\ref{fig:fluency-multisynt} reflects real differences rather than judge noise.

\section{Embedding-space coverage analysis}
\label{sec:appendix-infogap}

This appendix reports the embedding-space benchmark-alignment diagnostic summarized in Section~\ref{subsec:benchmark-alignment}.
We use it as a descriptive check on benchmark-adjacent coverage rather than as a causal test of the gains in Section~\ref{subsec:gains}.
The analysis measures which data source aligns more closely with benchmark regions in a shared embedding space.

\paragraph{Embedding space.}
We map each document into an embedding space using \texttt{intfloat/multilingual-e5-large-instruct} \citep{wang2024multilingual}, producing one vector per document.

\begin{figure}[ht]
    \centering
    \includegraphics[width=\linewidth]{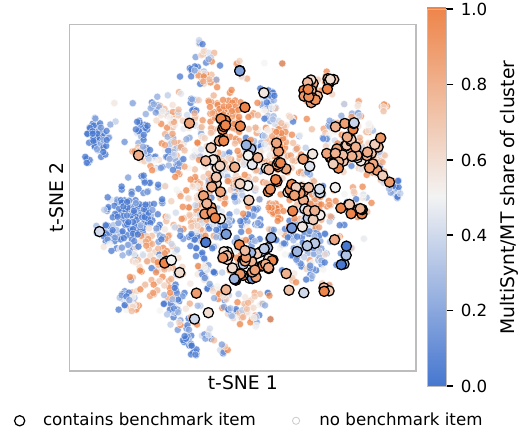}
    \caption{t-SNE projection of 2{,}000 k-means cluster centroids of the joint HPLT 2.0 + MultiSynt/MT pre-training embedding space for Spanish, colored by the per-cluster share of MultiSynt/MT documents (blue: HPLT 2.0-dominated; orange: MultiSynt/MT-dominated).
    Clusters that contain a Spanish ARC-Challenge benchmark item are outlined in black and tend to concentrate in MultiSynt/MT-denser regions of the space.}
    \label{fig:infogap-tsne}
\end{figure}

\begin{figure*}
    \centering
    \includegraphics[width=\linewidth]{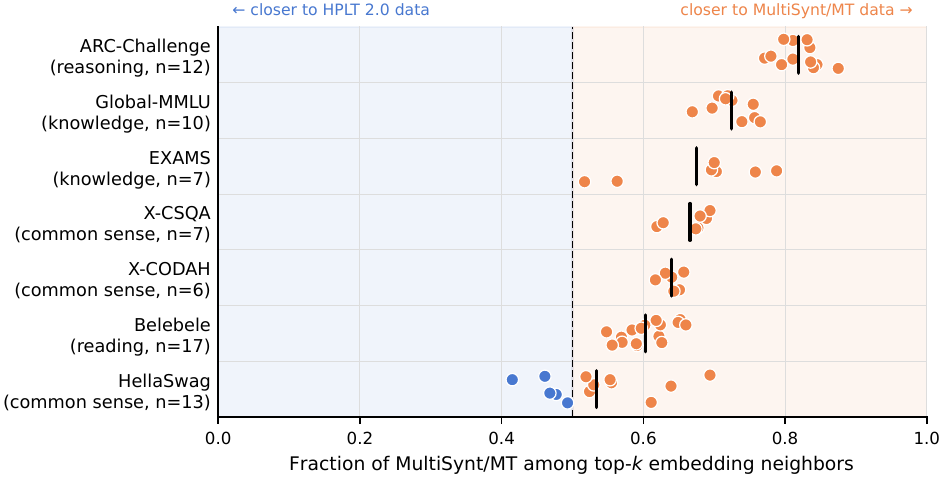}
    \caption{Embedding-space coverage analysis.
    For each benchmark, the fraction of MultiSynt/MT documents among the top-$k$ embedding neighbors of benchmark items is shown for every available language (one dot per language; $n$ in row labels).
    Black vertical ticks mark the per-benchmark mean across languages; the dashed line is parity at $0.5$.
    Six of seven benchmarks sit fully right of parity: their benchmark items have more MultiSynt/MT documents than HPLT 2.0 documents among their nearest embedding neighbors.
    HellaSwag is the lone exception, with five Latin-script languages on the HPLT 2.0 side.}
    \label{fig:infogap}
\end{figure*}

Figure~\ref{fig:infogap-tsne} illustrates benchmark-adjacent coverage for Spanish ARC-Challenge -- the benchmark that deviates most from parity in the main kNN figure (Figure~\ref{fig:infogap}) -- by partitioning the joint pre-training embedding space into $k=2{,}000$ clusters using k-means and assigning each benchmark example to its nearest cluster.
The two distributions partially overlap, but MultiSynt/MT covers several regions that are essentially unpopulated by HPLT 2.0 (and vice versa); clusters that contain a benchmark item tend to lie in MultiSynt/MT-denser regions, with a mean MultiSynt/MT share of $0.72$ versus $0.47$ for clusters that do not.
This is consistent with the per-language kNN ratio of $0.80$ for Spanish ARC-Challenge in Figure~\ref{fig:infogap}.

\paragraph{kNN coverage.}
To quantify this effect across benchmarks and languages, for each benchmark example, we retrieve its top-$k$ nearest neighbors from the union of native (HPLT 2.0) and synthetic (MultiSynt/MT) pre-training documents under cosine similarity, and report the average fraction $k_{\mathrm{synthetic}}/k$ across benchmark items for $k=20$.
A value of $0.5$ indicates equal local density from the two sources; values above $0.5$ indicate denser synthetic coverage of benchmark-adjacent regions.

The detailed results are shown in Figure~\ref{fig:infogap}.
Across all benchmarks, the mean values over the available languages indicate that the nearest documents are more often from MultiSynt/MT than from the native HPLT 2.0 data.
HellaSwag is the only benchmark for which some languages show a higher fraction of adjacent native documents, though its mean MultiSynt/MT fraction is still 0.53.
The other five benchmarks show a higher MultiSynt/MT fraction for all available languages.
ARC-Challenge deviates most from the parity line, with a MultiSynt/MT fraction of 0.82.
These results suggest that translated data increases the coverage of benchmark-adjacent regions, potentially bringing in content that is otherwise underrepresented in the native datasets.
However, we only conducted a descriptive study and did not account for causality of any potential gains.

\end{document}